\documentclass[pdflatex,sn-nature]{sn-jnl}

\usepackage{graphicx}
\usepackage{multirow}
\usepackage{amsmath,amssymb,amsfonts}
\usepackage{amsthm}
\usepackage{mathrsfs}
\usepackage[title]{appendix}
\usepackage{xcolor}

\usepackage{textcomp}
\usepackage{manyfoot}
\usepackage{booktabs}
\usepackage{algorithm}
\usepackage{algorithmicx}
\usepackage{algpseudocode}
\usepackage{listings}

\theoremstyle{thmstyleone}

\theoremstyle{thmstyletwo}

\theoremstyle{thmstylethree}

\begin{document}

\title[Ca2REAL]{Decoding spatial and semantic information from large-scale visual cortical activity for natural-movie reconstruction}

\author[1]{\fnm{Yukai} \sur{Huang}}\email{yukaihuang.research@outlook.com}
\equalcont{These authors contributed equally to this work.}

\author[1,2]{\fnm{Haoyu Albert} \sur{Wang}}\email{haoyuwang18@fudan.edu.cn}
\equalcont{These authors contributed equally to this work.}

\author[1]{\fnm{Jiawei} \sur{Zhang}}\email{Zhangjw22@m.fudan.edu.cn}
\equalcont{These authors contributed equally to this work.}

\author[1]{\fnm{Jialun} \sur{Ma}}\email{jlma26@m.fudan.edu.cn}
\author[1]{\fnm{Jiayao} \sur{Ding}}\email{jyding25@m.fudan.edu.cn}
\author[3]{\fnm{Yiran} \sur{Lin}}\email{yiran1.24@intl.zju.edu.cn}
\author[4]{\fnm{Xingyu} \sur{Chen}}\email{xingyuchen25@m.fudan.edu.cn}
\author[5]{\fnm{Ziqi} \sur{Gao}}\email{23307130412@m.fudan.edu.cn}
\author*[1]{\fnm{Yuguo} \sur{Yu}}\email{yuyuguo@fudan.edu.cn}

\affil*[1]{\orgdiv{MOE Frontiers Center for Brain Science, State Key Laboratory of Brain Function and Disorders, Research Institute of Intelligent and Complex Systems, and Institute of Science}, \orgname{Fudan University}, \orgaddress{\city{Shanghai}, \country{China}}}
\affil[2]{\orgdiv{Institute of Science and Technology for Brain-inspired Intelligence}, \orgname{Fudan University}, \orgaddress{\city{Shanghai}, \country{China}}}
\affil[3]{\orgdiv{University of Edinburgh Institute International Campus}, \orgname{Zhejiang University}, \orgaddress{\city{Hangzhou}, \country{China}}}
\affil[4]{\orgdiv{School of Life Sciences}, \orgname{Fudan University}, \orgaddress{\city{Shanghai}, \country{China}}}
\affil[5]{\orgdiv{College of Computer Science and Artificial Intelligence}, \orgname{Fudan University}, \orgaddress{\city{Shanghai}, \country{China}}}

\abstract{
Reconstructing natural visual experience from cortical population activity is a central challenge in neural decoding, particularly for calcium imaging, where signals are noisy, temporally filtered, and paired neural--video data are limited. Here we present \textbf{Ca2REAL} (\textbf{Calcium Activity to Reconstruct Events, Appearance, and Layout}), a neural-to-video framework that conditions a pretrained video diffusion model through separate layout and semantic controls derived from calcium activity. Using measured responses from 7,863 V1 neurons in one mouse, Ca2REAL reconstructed held-out natural-movie clips with 81.4\% all-pairs stimulus-identification accuracy, with its clearest gains over deterministic baselines in perceptual and semantic measures. We further evaluated the framework in a stimulus-aligned virtual population of 116,705 neurons assembled from encoding-model-predicted responses across 11 independently recorded datasets. Perturbation and ablation analyses showed that reconstruction depends on distributed, stimulus-specific population structure, with layout and semantic pathways providing complementary spatial and identity information. Temporal analyses further showed improved preservation of semantic dynamics across reconstructed clips. These results show that large-scale cortical population activity constrains the spatial, semantic, and temporal structure of natural visual experience, providing a generative framework for high-fidelity neural reconstruction of dynamic visual scenes.
}

\maketitle

\section*{Introduction}

The central goal of neural decoding is to infer information about external stimuli from the neural population responses they evoke, providing a quantitative approach for probing how sensory information is represented in the brain and an important foundation for brain--computer interfaces \citep{naselaris2011,kriegeskorte2019,wolpaw2002}. Vision has become a major testbed for neural decoding because natural scenes jointly contain rich spatial structure, visual identity, and temporal dynamics, making natural-movie reconstruction a stringent benchmark for recovering complex sensory information from brain activity. With advances in large-scale optical imaging, calcium imaging now enables cellular-resolution measurements from thousands of identified neurons during naturalistic visual stimulation \citep{devries2020,turishcheva2023,turishcheva2024}, creating new opportunities to study visual decoding at the scale of large neuronal populations.

Conventional deep-learning approaches reconstruct visual stimuli by learning mappings from neural activity to pixels or visual features. For functional magnetic resonance imaging (fMRI), responses can be mapped to hierarchical convolutional features, followed by feature inversion or learned image reconstruction to recover static images and individual movie frames \citep{shen2019,wen2018,zhang2018cnn}. At cellular resolution, the spike-image decoder (SID) combines a neural-to-image mapping with convolutional refinement \citep{zhang2020sid}. Other retinal and cortical methods use brain-inspired network designs and temporal representations, including wavelet-based spike features \citep{li2023braininspired,yu2025robust,peng2025wisa}. Calcium-based studies have likewise reconstructed image features or movie frames using population readouts and learned neural representations \citep{yoshida2020,zhang2022,bae2026}. However, when direct regression is trained with pixel-wise error objectives, uncertainty about the stimulus can favor averaged, blurred predictions, as illustrated by studies of video prediction and the perception--distortion tradeoff \citep{mathieu2016,blau2018}. These limitations are compounded in calcium imaging: indicator kinetics temporally filter neural activity, measurement noise reduces response reliability, and fluorescence signals provide only an indirect readout of underlying spiking activity \citep{chen2013gcamp,theis2016,deneux2016,rupprecht2021}. With limited paired neural--video data, conventional supervised decoders must therefore learn a high-dimensional inverse mapping from noisy, temporally blurred inputs, making simultaneous recovery of fine visual detail and temporal structure particularly difficult.

Recent advances in diffusion models offer new opportunities for neural reconstruction. These models learn an iterative denoising process that captures natural-image and video statistics and supports generation conditioned on external representations \citep{rombach2022,ho2022video}. In fMRI decoding, conditioning pretrained diffusion models on neural-derived visual and semantic features has enabled high-resolution natural-image reconstruction \citep{mindvis2023,takagi2023,ozcelik2023}, while related approaches have combined semantic information with temporal constraints for video reconstruction \citep{chen2023,lu2024,gong2024,wang2025neurons}. Such models can supply visual detail from a learned prior, reducing the need to infer every pixel directly from paired neural data. Diffusion-based reconstruction from calcium activity has also been demonstrated for static images \citep{yang2024,deng2026}. Extending these advances to natural-movie reconstruction from calcium population activity remains particularly challenging because the decoder must recover coherent visual content and its temporal evolution from noisy, temporally filtered, and incomplete neural signals. Pretrained generative models provide a powerful prior, but they also introduce an attribution problem: realistic content may be generated by the prior rather than constrained by the recorded calcium activity. The key challenge is therefore to combine generative priors with reliable, stimulus-specific constraints derived from cortical calcium responses.

Here, we present \textbf{Ca2REAL} (\textbf{Calcium Activity to Reconstruct Events, Appearance, and Layout}), a generative neural-to-video framework that conditions a pretrained video diffusion model on complementary layout and semantic constraints derived from calcium activity. We first evaluated Ca2REAL using measured responses from 7,863 V1 neurons recorded in a single mouse. On 178 held-out two-second natural-movie clips, Ca2REAL achieved 81.4\% all-pairs stimulus-identification accuracy, with its clearest advantages over deterministic decoders in perceptual and semantic correspondence. We then examined decoding with a stimulus-aligned virtual population of 116,705 neurons assembled from encoding-model-predicted responses across 11 independently recorded datasets spanning V1, LM, RL, and AL, where Ca2REAL reached 92.8\% all-pairs stimulus-identification accuracy. Measured-response substitution further tested the functional compatibility of experimentally recorded responses with the virtual-population decoder. Population perturbations showed that reconstruction depends on distributed, structured, and stimulus-specific neural information, while pathway ablations revealed complementary contributions of layout and semantic conditioning to spatial organization and visual identity. Temporal analyses further showed improved preservation of visual-semantic dynamics across reconstructed clips. Together, Ca2REAL establishes a scalable and interpretable approach for reconstructing natural movies from cellular-resolution calcium activity and for systematically testing how population-level spatial and semantic information shapes generative visual reconstruction.

\begin{figure}[!t]
\centering
\includegraphics[width=\textwidth]{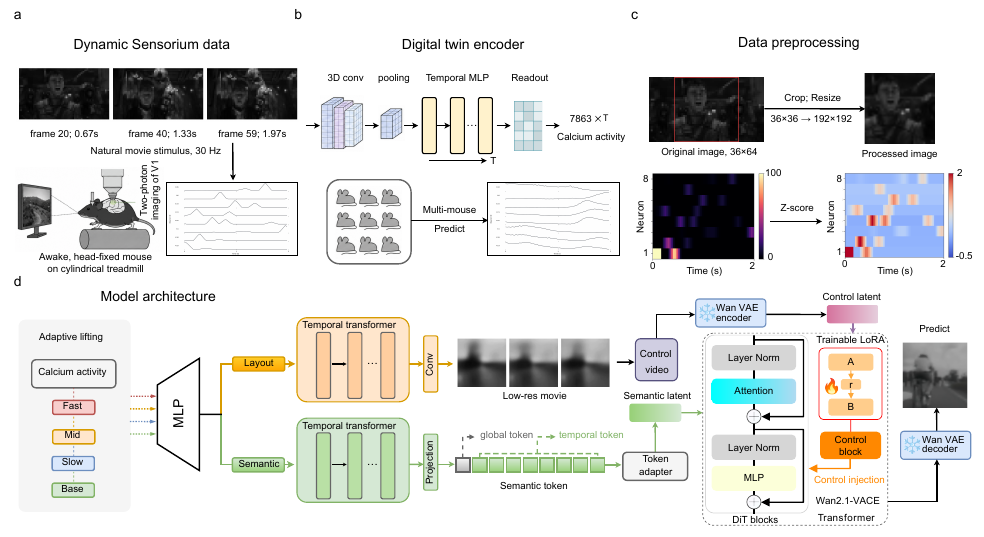}
\caption{\textbf{Overview of the data pipeline and Ca2REAL architecture.} \textbf{a}, Dynamic Sensorium data acquisition for the primary measured experiment. Awake, head-fixed mice viewed natural movies presented at 30~Hz while locomoting on a cylindrical treadmill, and V1 calcium activity was recorded with two-photon imaging. \textbf{b}, Neural-response synthesis with digital-twin encoding models. A spatiotemporal neural encoder maps each movie to the corresponding population response through a shared visual feature extractor, temporal processing, and population-specific readout. \textbf{c}, Neural and visual preprocessing. Movie frames were centrally cropped and resized to the generator resolution, while calcium responses were normalized neuron-wise using training-set statistics before decoding. \textbf{d}, Ca2REAL architecture. A multiscale temporal front end provides features to two neural-control pathways. The layout pathway predicts a low-resolution movie for spatial control of Wan2.1-VACE, whereas the semantic pathway predicts global and temporally ordered visual tokens that are adapted to the generator conditioning space.}
\label{fig:ca2real-overview}
\end{figure}

\section*{Results}

\subsection*{Experimental datasets and Ca2REAL framework}

We evaluated Ca2REAL in two neural-data settings that shared a common preprocessing and decoding pipeline (Fig.~\ref{fig:ca2real-overview}). The primary experiment used Dynamic Sensorium recordings \citep{turishcheva2023,turishcheva2024}, in which awake, head-fixed mice viewed natural movies while V1 calcium activity was recorded with two-photon imaging. We analyzed responses from 7,863 neurons in one mouse using two-second, 60-frame neural--video segments. To examine whether the decoding framework remains effective when scaled to substantially larger neural populations than are currently available in a single simultaneous recording, we additionally constructed a stimulus-aligned virtual population. This population combined ten Sensorium V1 datasets with a MICrONS-derived population spanning V1, LM, RL, and AL \citep{microns2025}. Dataset-specific encoding models predicted responses to a common set of movie stimuli, and the stimulus-aligned predictions were concatenated across populations to yield a 116,705-neuron virtual population. This construction therefore provides a controlled large-scale decoding setting rather than a simultaneous biological recording.

Ca2REAL was designed to extract complementary spatial, semantic, and temporal information from calcium population activity while using a pretrained video prior to recover visual detail that is incompletely specified by the neural signal. Neural activity first passes through a multiscale temporal front end that decomposes the calcium sequence into features with different temporal dynamics, providing a common representation for two specialized neural-control pathways. The layout pathway predicts a low-resolution spatiotemporal scaffold that constrains coarse scene geometry, object location, luminance structure, and their evolution over time. In parallel, the semantic pathway predicts both clip-level and temporally ordered visual representations, providing stimulus-specific information about visual identity and changes in scene content. The two neural-derived controls are then integrated by a pretrained Wan2.1-VACE video generator \citep{wan2025,jiang2025}, which supplies learned natural-video regularities within these spatial and semantic constraints. This separation is central to the design of Ca2REAL: rather than asking a deterministic decoder to predict every visual detail directly from noisy calcium activity, the model assigns coarse organization and stimulus identity to explicit neural-derived controls while using the pretrained generator for prior-supported visual completion. Detailed acquisition, preprocessing, architecture, and training procedures are provided in Methods.

\begin{figure}[!hbpt]
\centering
\includegraphics[width=\textwidth]{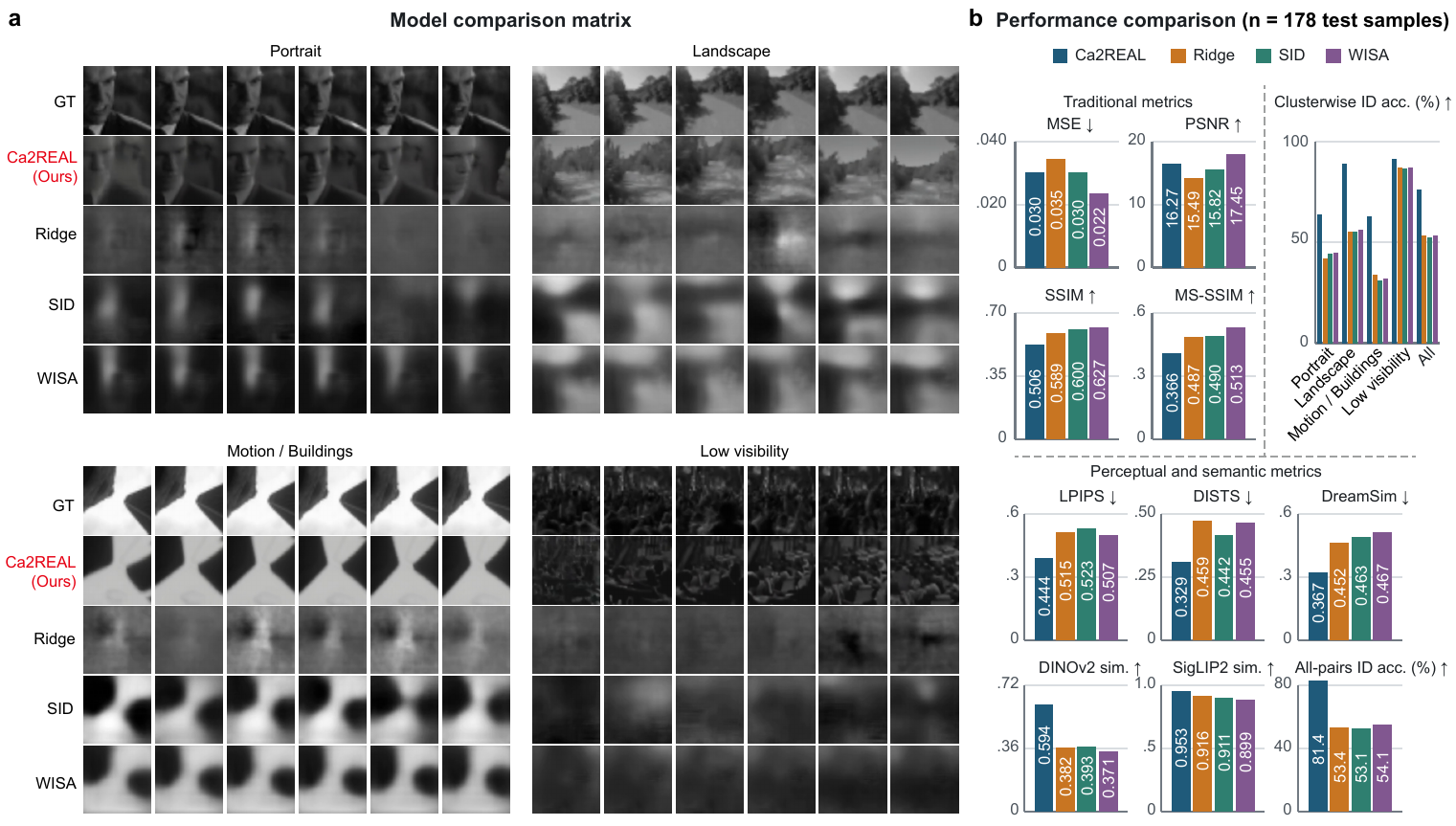}
\caption{\textbf{Natural-movie reconstruction from measured V1 calcium activity in a single mouse.} \textbf{a}, Representative held-out target clips (GT) and reconstructions generated by Ca2REAL and baseline methods from measured responses of 7,863 V1 neurons. Selected frames from one representative clip in each of the four target clusters are shown. \textbf{b}, Quantitative reconstruction performance on 178 held-out measured clips.}
\label{fig:temporary-evaluation-sketch}
\end{figure}

\subsection*{Ca2REAL reconstructs natural movies from measured V1 activity}

In the measured-response experiment, reconstruction was assessed on 178 held-out two-second natural-movie clips using experimentally measured V1 activity. Ca2REAL produced stimulus-specific reconstructions that preserved recognizable scene structure and visual content across time (Fig.~\ref{fig:temporary-evaluation-sketch}). Representative examples are shown in Fig.~\ref{fig:temporary-evaluation-sketch}a for the four post hoc target clusters: \emph{Portrait}, \emph{Landscape}, \emph{Motion / Buildings}, and \emph{Low visibility}. In the displayed \emph{Portrait} example, Ca2REAL preserved recognizable facial structure that was weak in the deterministic reconstructions. The \emph{Landscape} example retained the large-scale arrangement of sky, road, vegetation, and terrain, while the \emph{Motion / Buildings} example preserved major structural boundaries across the sampled frames. Under \emph{Low visibility}, Ca2REAL retained more identifiable local structure than the smoother baseline outputs. Across all four categories, the reconstructed content remained temporally consistent from frame to frame, preserving scene identity and spatial organization throughout the video sequence.

Quantitative evaluation revealed a distinction between pixel-wise fidelity and perceptual reconstruction quality (Fig.~\ref{fig:temporary-evaluation-sketch}b). Ca2REAL showed no systematic advantage on conventional pixel-level measures such as MSE, PSNR, SSIM, and MS-SSIM, which directly reward pointwise agreement with the target and are closely matched to the objectives of deterministic regression-based decoders. In contrast, Ca2REAL showed large and consistent gains on measures designed to capture perceptual similarity and semantic content, achieving the best LPIPS ($0.444$), DISTS ($0.329$), DreamSim ($0.367$), DINOv2 similarity ($0.594$), and SigLIP2 similarity ($0.953$). This metric profile is consistent with the qualitative reconstructions: the deterministic baselines often remained closer in a pixel-averaged sense but produced smoother predictions, whereas Ca2REAL better preserved stimulus-specific structure and recognizable visual content. All-pairs stimulus-identification accuracy provided a complementary test of target specificity, asking whether each reconstruction was closer in semantic feature space to its own target than to mismatched alternatives. Ca2REAL ranked the matched target above a mismatched target in $81.4\%$ of ordered comparisons, compared with $54.1\%$ for the strongest baseline, and the advantage was present in each of the four target clusters. Together, these results show that measured V1 population activity supports reconstruction that preserves stimulus-specific perceptual and semantic information even when exact pixel-level correspondence is incomplete.

\subsection*{Natural-movie decoding from a 116,705-neuron virtual population}

\begin{figure}[!t]
\centering
\includegraphics[width=\textwidth]{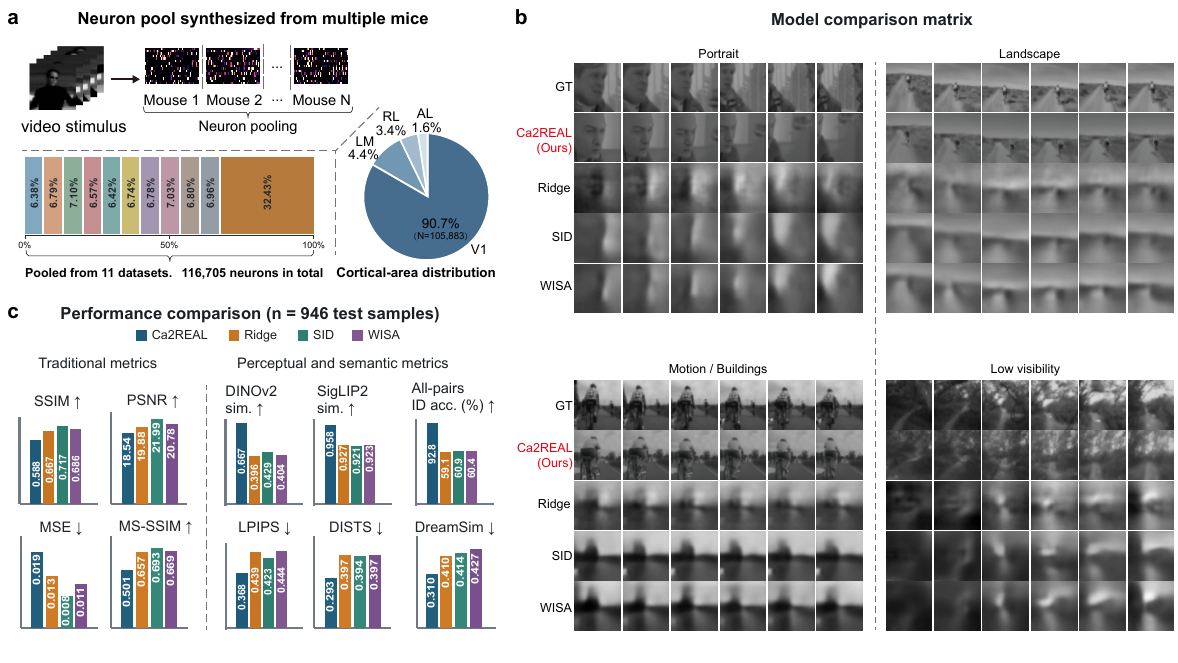}
\caption{\textbf{Neural-to-video decoding from a stimulus-aligned virtual population of 116,705 neurons.} \textbf{a}, Construction of the virtual population. Encoding models fitted to 11 independently recorded neural datasets predicted responses to common movie stimuli, and the predictions were concatenated across neurons. The lower panels show the contribution of each dataset and the cortical-area composition of the virtual population. \textbf{b}, Representative held-out target clips (GT) and reconstructions generated by Ca2REAL and baseline methods from the 116,705-dimensional encoding-model-predicted responses. Selected frames from one representative clip in each target cluster are shown. \textbf{c}, Quantitative comparison over 946 held-out segments.}
\label{fig:temporary-digital-twin-sketch}
\end{figure}

The 116,705-neuron virtual population was dominated by V1 neurons (90.7\%, $N=105{,}883$), with smaller contributions from LM (4.4\%, $N=5{,}080$), RL (3.4\%, $N=3{,}912$), and AL (1.6\%, $N=1{,}830$) (Fig.~\ref{fig:temporary-digital-twin-sketch}a). This area composition reflects pooled encoding-model predictions from independently recorded source populations and should not be interpreted as a simultaneous multi-area calcium recording.

Ca2REAL was then trained on the 116,705-dimensional virtual responses. Representative reconstructions showed the same qualitative pattern as in the measured-data analysis: Ca2REAL preserved more recognizable stimulus content and scene structure, whereas the deterministic outputs were generally smoother (Fig.~\ref{fig:temporary-digital-twin-sketch}b). This qualitative difference was accompanied by strong perceptual, semantic, and identification performance across the 946 held-out segments. Ca2REAL achieved SSIM $0.588$, LPIPS $0.368$, DINOv2 similarity $0.667$, SigLIP2 similarity $0.958$, and $92.8\%$ all-pairs stimulus-identification accuracy (Fig.~\ref{fig:temporary-digital-twin-sketch}c). Thus, in $92.8\%$ of ordered matched-versus-mismatched comparisons, the reconstructed clip was closer in SigLIP2 space to its own target, compared with identification accuracies of $59.1\%$ for Ridge, $60.9\%$ for SID, and $60.4\%$ for WISA. These results show that Ca2REAL can reconstruct held-out movie segments from the 116,705-dimensional virtual responses while retaining substantially stronger stimulus-specific correspondence than the deterministic baselines.

\begin{table}[!t]
\centering
\caption{\textbf{Measured-response substitution tests functional compatibility with the virtual-population decoder.}
All-pairs identification accuracy after replacing available synthetic responses with matched measured activity or control responses on the full test set ($n=946$ clips). $\Delta$ denotes the change relative to the fully synthetic condition.}
\label{tab:measured-substitution}
\small
\begin{tabular*}{\textwidth}{@{\extracolsep{\fill}}lcc@{}}
\hline
\textbf{Condition} & \textbf{ID Acc. (\%)} & \textbf{$\Delta$ (pp)} \\
\hline
Synthetic & 92.83 & 0.00 \\
Matched measured & 91.98 & $-0.85$ \\
Shuffled measured & 91.10 & $-1.73$ \\
Zeroed responses & 91.46 & $-1.37$ \\
\hline
\end{tabular*}
\end{table}

We next asked whether available experimentally measured responses could substitute for their encoding-model predictions within the same virtual-population decoder. Approximately 15,000 neurons per clip (about 13\% of the 116,705-dimensional population) had a measured counterpart. We compared matched measured substitution with controls in which the same replaceable channels were shuffled across stimuli or set to zero. All-pairs identification decreased from 92.83\% for the fully synthetic input to 91.98\% with matched measured substitution, compared with 91.10\% for shuffled responses and 91.46\% for zeroed responses (Table~\ref{tab:measured-substitution}). Matched measured responses therefore produced the smallest degradation among these substitution conditions. Because measured substitution covered only a minority of channels and did not improve performance over the fully synthetic input, this result supports functional compatibility of the available measured responses with the decoder rather than validating the virtual population as a biological recording.

\begin{figure}[!h]
\centering
\includegraphics[width=\textwidth]{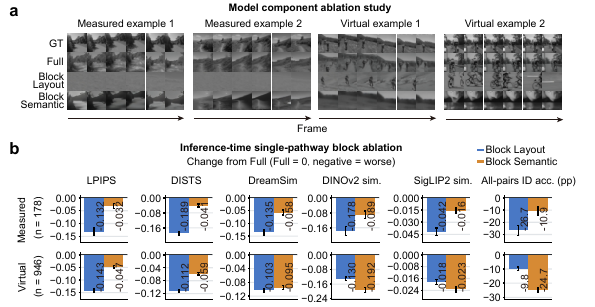}
\caption{\textbf{Complementary roles of the layout and semantic conditioning pathways.} \textbf{a}, Representative reconstructions from measured and virtual-population activity using the full Ca2REAL model or with the layout or semantic pathway blocked at inference time. \textbf{b}, Mean oriented paired change in reconstruction performance relative to the full model after blocking either pathway. Signs are oriented so that negative values indicate worse performance for both higher-is-better and lower-is-better metrics. Results are shown separately for the measured V1 population ($n=178$ clips) and the virtual population ($n=946$ clips).}
\label{fig:temporary-ablation-sketch}
\end{figure}

\subsection*{Layout and semantic pathways provide complementary reconstruction constraints}

To determine how the two conditioning pathways contribute to reconstruction, we blocked either the layout or semantic control at inference while leaving model weights, the remaining control, and diffusion sampling unchanged (Fig.~\ref{fig:temporary-ablation-sketch}). The two interventions produced distinct failure modes (Fig.~\ref{fig:temporary-ablation-sketch}a). Without layout control, reconstructions often lost the spatial organization of the target, whereas without semantic control, coarse geometry was more often retained while visual identity changed. The semantic effect was particularly evident in stimulus-specific details: characteristic forest and grassland textures were largely lost in the measured V1 examples, while the cyclists disappeared from both virtual-population examples despite broadly preserved spatial organization; in one of these reconstructions, the generated content instead included an unrelated face-like pattern. These qualitative differences suggest that the two pathways constrain different aspects of the reconstructed movie rather than providing interchangeable conditioning signals.

The paired changes show the same dissociation (Fig.~\ref{fig:temporary-ablation-sketch}b). Changes are oriented so that negative values denote worse performance for every metric, including lower-is-better distances. In the measured data, blocking layout caused larger losses than blocking semantics for LPIPS ($-0.132$ vs. $-0.032$), DISTS ($-0.189$ vs. $-0.041$), and all-pairs identification accuracy ($-26.7$ vs. $-10.9$ percentage points). In the virtual population, layout blocking produced the larger LPIPS loss ($-0.143$ vs. $-0.047$), whereas semantic blocking produced the larger DINOv2 loss ($-0.192$ vs. $-0.130$) and identification loss ($-24.7$ vs. $-9.8$ percentage points). Within the trained architecture, layout control therefore contributes more strongly to spatial and perceptual fidelity, while semantic control contributes more strongly to stimulus identity in the virtual-population analysis.

\subsection*{Reconstruction depends on distributed and stimulus-specific neural population structure} 

\begin{figure}[!t]
 \centering
 \includegraphics[width=\textwidth]{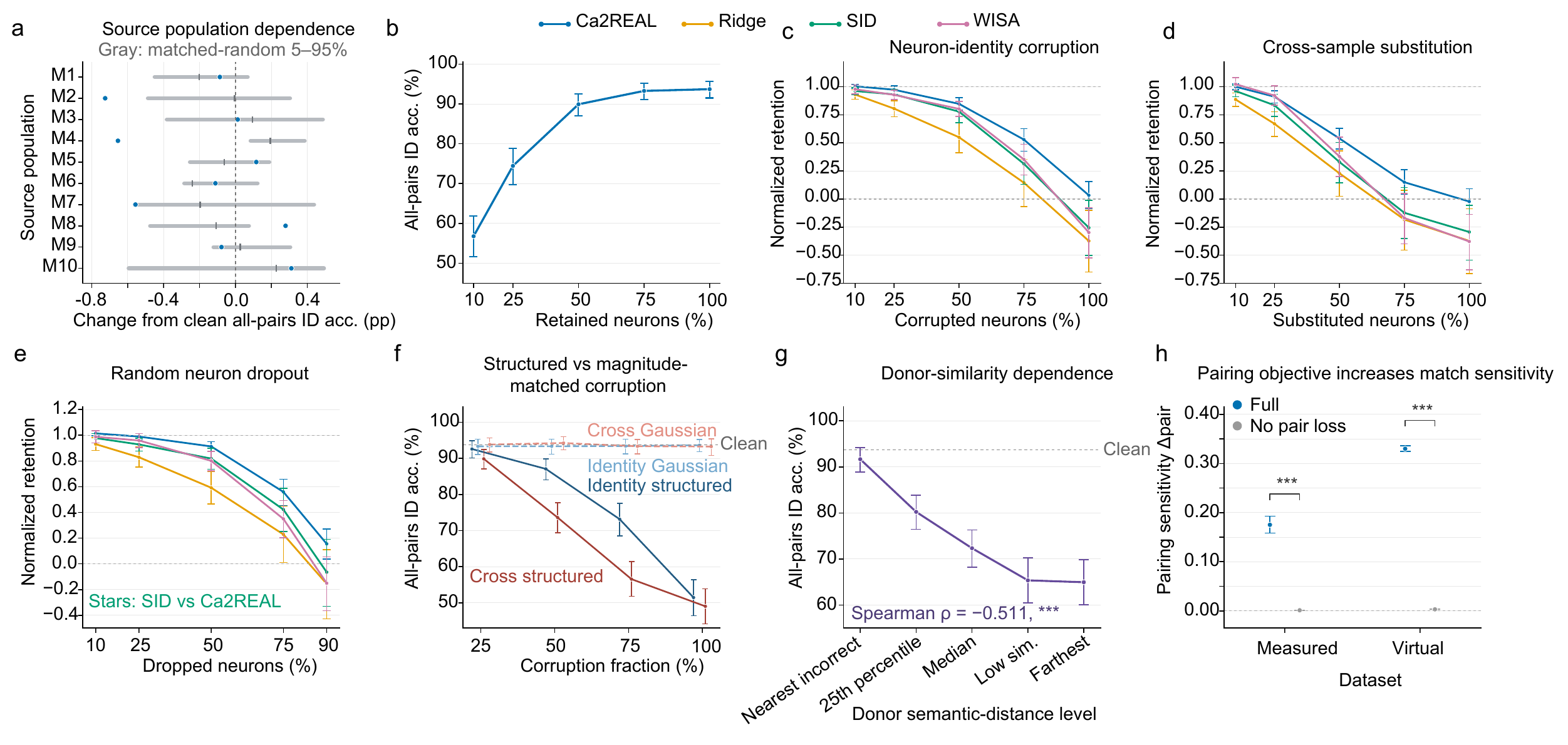}
 \caption{\textbf{Distributed and stimulus-specific neural dependence of Ca2REAL.} \textbf{a}, Effect of removing each source population relative to matched-size random neuron deletions (blue point, source-population removal result; gray line, matched-random 5--95\% range; gray tick, median). \textbf{b}, Reconstruction performance as a function of the fraction of retained neurons.
\textbf{c}, Robustness to progressive corruption of neuron identity.
\textbf{d}, Robustness to replacing neural responses with responses evoked by a different stimulus.
\textbf{e}, Robustness to random neuron dropout.
\textbf{f}, Structured neuron-identity and cross-sample perturbations compared with magnitude-matched Gaussian perturbations.
\textbf{g}, Effect of cross-sample substitution as a function of target--donor similarity defined in frozen SigLIP2 space; the outcome is SigLIP2-based all-pairs identification, and significance was assessed by a cluster-aware permutation test of the global Spearman trend.
\textbf{h}, Denoising-objective pairing sensitivity at fixed diffusion noise level ($\sigma=0.7$) in the full model and the model trained without $\mathcal{L}_{\mathrm{pair}}$,
quantified by
$\Delta_{\mathrm{pair}}
=\mathcal{L}_{\mathrm{shuffled}}
-\mathcal{L}_{\mathrm{matched}}$;
asterisks indicate paired Full versus
No-$\mathcal{L}_{\mathrm{pair}}$ comparisons.
For \textbf{b--h}, vertical bars indicate bootstrap 95\% confidence intervals.
}
 \label{fig:distributed-neural-dependence}
\end{figure}

We first asked whether reconstruction depended disproportionately on individual source populations. For each of the ten source populations tested in Fig.~\ref{fig:distributed-neural-dependence}a, we removed all neurons from that population and compared the resulting change in all-pairs identification accuracy with matched-size random neuron removals. Removing any of these tested sources changed accuracy by less than one percentage point, with no tested source producing a marked loss in performance. In contrast, progressively reducing the available population produced a systematic decline in decoding (Fig.~\ref{fig:distributed-neural-dependence}b). Accuracy remained near the full-population level at 75\% retention, decreased by 3.79 percentage points at 50\% retention, and fell to 56.79\% when only 10\% of neurons were retained. Thus, the population-size analysis supports distributed dependence within the tested decoder.

We then examined whether this distributed information depended on the structure of the population response. We applied neuron-identity corruption (assigning responses to incorrect neuron identities), cross-sample substitution (replacing responses with those evoked by another stimulus), and random neuron dropout (randomly setting neuronal activity to zero) (Fig.~\ref{fig:distributed-neural-dependence}c--e). Across most perturbation levels, Ca2REAL degraded more slowly than Ridge, SID, and WISA, indicating greater robustness to incomplete or corrupted neural input. The effect within Ca2REAL nevertheless depended strongly on the form of perturbation: cross-sample substitution consistently produced the largest loss, indicating that reconstruction relies on correspondence between the observed population response and the current stimulus. This sensitivity was not explained by perturbation magnitude. Magnitude-matched Gaussian perturbations left decoding close to the clean level, whereas complete structured perturbations reduced all-pairs identification by more than 42 percentage points relative to their Gaussian controls (Fig.~\ref{fig:distributed-neural-dependence}f). The resulting deficit therefore reflects disruption of structured neural information rather than nonspecific changes in response magnitude.

We next asked whether sensitivity to cross-sample substitution reflected graded stimulus structure in the population response. At 50\% substitution, we selected donor stimuli spanning different levels of similarity to the target in a frozen SigLIP2 representation (Fig.~\ref{fig:distributed-neural-dependence}g). All-pairs identification was $91.66\%$ when the substituted responses were evoked by the nearest incorrect semantic neighbor and declined progressively to $64.96\%$ for the most distant donor. Consistently, greater target--donor similarity was associated with less substitution-induced degradation (Spearman $\rho=-0.511$, cluster-aware permutation $p=1.0\times10^{-5}$). Thus, neural responses evoked by similar visual content were more compatible with reconstruction of the same target, indicating that Ca2REAL captures graded stimulus-related structure in population activity rather than simply associating each neural response pattern with a particular video.

Finally, we examined whether this sensitivity to neural--stimulus correspondence was promoted by the matched-versus-shuffled dependence loss introduced during training (Eq.~\ref{eq:pair-dependence}). At a fixed diffusion noise level ($\sigma=0.7$), we quantified matching sensitivity as $\Delta_{\mathrm{pair}}=\mathcal{L}_{\mathrm{shuffled}}-\mathcal{L}_{\mathrm{matched}}$ and compared the full model with a model trained without $\mathcal{L}_{\mathrm{pair}}$ (Fig.~\ref{fig:distributed-neural-dependence}h). In the measured population, $\Delta_{\mathrm{pair}}$ decreased from $0.175$ to $0.0013$ after removing $\mathcal{L}_{\mathrm{pair}}$; in the virtual population, it decreased from $0.330$ to $0.0032$. Removing the pairing objective therefore nearly abolished the denoising objective's sensitivity to whether neural activity was correctly matched to the stimulus. Together, these results show that Ca2REAL draws on distributed population information while preserving neuron identity and graded stimulus-specific structure, with the pairing objective providing the training signal that reinforces this neural--stimulus correspondence.

\subsection*{Ca2REAL preserves temporal semantic dynamics} 

\begin{figure}[!h]
 \centering
 \includegraphics[width=\textwidth]{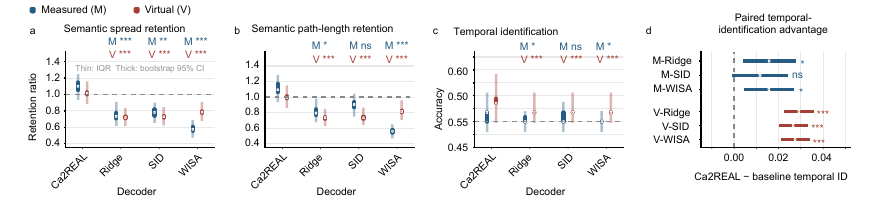}
\caption{
\textbf{Ca2REAL preserves temporal semantic dynamics.}
\textbf{a}, Semantic trajectory spread retention (points, medians; thin bars, interquartile ranges; thick bars, bootstrap 95\% CIs; dashed line, ideal ratio $=1$).
\textbf{b}, Semantic path-length retention, with conventions as in \textbf{a}.
\textbf{c}, Within-clip temporal identification accuracy (dashed line, chance level $=0.5$).
\textbf{d}, Paired temporal-identification advantage of Ca2REAL over each baseline (points, mean paired differences; bars, bootstrap 95\% CIs; dashed line, zero).
Blue/red denote measured/virtual populations. Significance tests were Holm-corrected; asterisks denote significance and ns denotes non-significance.
}
 \label{fig:temporal-semantic-dynamics}
\end{figure}

To test whether reconstructed movies preserved temporal semantic organization, we embedded eight fixed time points from each target and reconstruction using the same frozen semantic encoder. The resulting trajectories were evaluated by their semantic spread, accumulated path length, and within-clip temporal correspondence (Fig.~\ref{fig:temporal-semantic-dynamics}a--c), with paired differences between Ca2REAL and each baseline summarized in Fig.~\ref{fig:temporal-semantic-dynamics}d. Spread and path retention were normalized to the corresponding target trajectory, such that a ratio of 1 indicates perfect retention, whereas temporal-identification accuracy has a chance level of 0.5.

Across complementary measures of semantic trajectory geometry and temporal alignment, Ca2REAL preserved temporal semantic dynamics more faithfully than the deterministic baselines (Fig.~\ref{fig:temporal-semantic-dynamics}a--d). Across both populations, Ca2REAL preserved semantic spread close to the target scale, with median retention ratios of 1.102 and 1.015 in the measured and virtual populations, respectively, compared with 0.733/0.719 for Ridge, 0.782/0.728 for SID, and 0.575/0.784 for WISA (Fig.~\ref{fig:temporal-semantic-dynamics}a). Using paired sign-flip tests on the absolute deviation from the ideal ratio of 1, Ca2REAL was significantly closer to the target spread than all three baselines in both populations after Holm correction. Path-length retention showed the same overall trend (Fig.~\ref{fig:temporal-semantic-dynamics}b): Ca2REAL remained close to the target in both measured and virtual populations (1.094 and 0.997), whereas the baselines were generally more contracted. In the measured population, Ca2REAL was significantly closer to the target path length than Ridge and WISA but not SID; in the virtual population, it was significantly closer than all three baselines. Ca2REAL also improved within-clip temporal identification (Fig.~\ref{fig:temporal-semantic-dynamics}c,d), increasing the median accuracy from 0.500 to 0.518 in the measured population and from 0.518 to 0.536 in the virtual population; the paired advantage was significant over Ridge and WISA, but not SID, in the measured population and over all three baselines in the virtual population. Together, these measures show that the advantage extends beyond frame-wise semantic similarity: Ca2REAL better preserves both the scale of semantic trajectories and the correspondence of semantic states across time.

\section*{Discussion}

Our results indicate that cortical calcium population activity provides information that constrains scene organization, visual identity, and aspects of their temporal evolution during natural-movie reconstruction. The measured-response experiment establishes that these properties are recoverable from recorded activity. In the virtual-population analysis, the small effects of removing individual tested source populations, together with the decline under progressive neuron removal, support distributed dependence within the tested decoder (Fig.~\ref{fig:distributed-neural-dependence}). The greater effects of neuron-identity corruption and cross-sample substitution than of magnitude-matched Gaussian perturbations further indicate that reconstruction depends on structured, stimulus-specific response patterns rather than response magnitude alone. Previous studies have linked natural-scene discriminability to sparse population responses \citep{froudarakis2014} and described high-dimensional response geometries in visual cortex \citep{stringer2019}. Our analyses complement these descriptions by connecting the organization of decoder inputs to the visual content that can be recovered from them.
These findings extend earlier calcium-based reconstruction studies using conventional or diffusion-based image decoders \citep{yoshida2020,zhang2022,yang2024,deng2026} and recent approaches to dynamic calcium decoding based on encoding-model inversion or deterministic reconstruction \citep{bauer2026,bae2026}.
The contribution is therefore both a reconstruction framework and a way to test which properties of population responses constrain naturalistic visual readout.

The use of a generative prior changes how this recoverable information should be interpreted.
The layout pathway provides a coarse spatiotemporal constraint, whereas the semantic pathway supplies stimulus-specific visual information; the pretrained video backbone can then contribute high-frequency texture and object detail that are only weakly specified by calcium activity. Deterministic decoders can perform well on MSE, PSNR, or SSIM by producing smooth predictions that remain close to the target in an averaged pixel sense, whereas Ca2REAL trades some of this pixel-level agreement for stronger perceptual and semantic correspondence.
This divergence is consistent with the perception--distortion tradeoff, which distinguishes low reconstruction error from perceptual realism \citep{blau2018}, and complements generative fMRI-to-video studies that use pretrained visual priors to recover naturalistic content \citep{chen2023,lu2024,gong2024}.
Such prior-supported completion is useful only if the neural signal still determines meaningful aspects of the generated scene.
The pathway ablations, structured perturbations, and cross-sample substitutions support this dependence, while the matched-versus-shuffled analysis shows that the training objective reinforces sensitivity to neural--stimulus correspondence.

The pathway and temporal analyses further clarify what aspects of the movie are constrained by the neural response. Layout conditioning contributed more strongly to spatial and perceptual organization, whereas semantic conditioning contributed more strongly to stimulus-specific visual identity, particularly in the virtual-population analysis. Their complementary effects suggest that coarse scene organization and higher-level visual content can be read out through partially separable representations rather than a single undifferentiated neural condition.
These analyses remain descriptions of information used by the trained decoder rather than direct evidence for distinct biological pathways, but they provide a controlled way to relate properties of large neural populations to complementary spatial, semantic, and temporal constraints on natural-movie reconstruction.  Temporal analyses further showed preservation of semantic-trajectory geometry and within-clip correspondence, with the strength of the advantage varying across comparisons (Fig.~\ref{fig:temporal-semantic-dynamics}). This connects our reconstruction results to prior work showing structured trajectories of natural-video representations in macaque V1 \citep{henaff2021} and stable readout of movie time from variable mouse V1 activity \citep{xia2021}. Preservation of these semantic trajectories therefore indicates that changes in visual content over time remain recoverable from population activity through the trained decoder.

The 116,705-neuron virtual-population experiment primarily tests the scaling behavior of the decoding framework in an encoding-model-mediated setting. Calcium-based movie reconstruction also remains underdetermined at fine spatial and temporal scales, so further progress will depend on separating information directly supported by neural activity from structure supplied by increasingly powerful generative priors. Input perturbations and pathway ablations establish dependence on neural information at the level of the tested controls, but cannot attribute every generated object, texture, or motion detail to a measured response. Comparisons across pretrained priors and experiments with controlled changes in visual content could sharpen this attribution. More broadly, Ca2REAL positions generative reconstruction as a controlled computational readout of population activity. With the visual prior held fixed, selective perturbations can reveal which response components support recoverable scene structure, identity, and temporal continuity. This approach complements conventional encoding and discrimination analyses \citep{naselaris2011,kriegeskorte2019} by making changes in recoverable content explicit, and could help test how that content varies across neuronal subsets, behavioral states, and stimulus conditions. Its broader value is to connect increasingly large neural recordings to testable questions about the organization of visual information, while maintaining an explicit distinction between neurally supported content and prior-driven completion.

\section*{Methods}

\subsection*{Neural data and preprocessing}

\noindent\textbf{Measured V1 dataset.}
The experimental and computational data pipeline is summarized in Fig.~\ref{fig:ca2real-overview}a--c. In the Dynamic Sensorium paradigm \citep{turishcheva2023,turishcheva2024}, awake, head-fixed mice viewed natural movies while running on a cylindrical treadmill, and V1 calcium activity was recorded with two-photon imaging (Fig.~\ref{fig:ca2real-overview}a). Natural movies were presented at 30~Hz. The primary measured experiment used 7,863 V1 neurons from one mouse. Each decoding example contained a two-second, 60-frame movie segment and the aligned neural responses. Most measured clips corresponded to a single stimulus presentation: 1,565 of 1,602 training clips and 175 of 178 test clips. For stimuli with repeated presentations, responses were averaged across the available repeats before normalization to ensure that each clip was represented the same number of times. The measured split contained 1,602 training clips and 178 held-out test clips.

\noindent\textbf{Digital-twin augmentation.}
Because paired measured calcium--video data are limited, neural encoding models were used only to augment the training set (Fig.~\ref{fig:ca2real-overview}b). Each digital twin maps a natural movie through a spatiotemporal visual encoder and population-specific readout to predict the responses of the recorded neurons \citep{deng2024encoding}. For the 7,863-neuron experiment, four independently trained encoding models predicted responses to stimuli already present in the measured training set, and their predictions were averaged. The resulting 1,443 synthetic-response entries were combined with 1,602 measured-response entries, giving 3,045 training examples. Evaluation used only the 178 experimentally measured responses.

\noindent\textbf{Stimulus-aligned virtual population.}
We separately constructed a stimulus-aligned virtual population to examine decoding with a substantially larger neural input. The ten Sensorium source populations were obtained by two-photon calcium imaging of V1 in awake, head-fixed mice during dynamic visual stimulation \citep{turishcheva2023,turishcheva2024}. The MICrONS source data were obtained by two-photon calcium imaging in an awake, head-restrained mouse and spanned V1 and the higher visual areas LM, RL, and AL during naturalistic and parametric visual stimulation \citep{microns2025}. Encoding models fitted to the ten Sensorium populations containing 78,853 neurons and a conservatively selected MICrONS-derived population containing 37,852 neurons were used to predict responses to the same movie stimuli. Their predicted responses were aligned by stimulus and concatenated across neurons, yielding a 116,705-dimensional virtual population. The constituent cells originate from independently recorded datasets, but the 116,705 responses used here are encoding-model predictions to common stimuli; they are not a simultaneous biological recording. The virtual-population experiment used 1,892 canonical 300-frame movies, each divided into five non-overlapping 60-frame segments, giving 9,460 neural--video pairs. A fixed segment-level split assigned 8,514 clips to training and 946 to testing, with identical memberships for Ca2REAL and all baselines.

\noindent\textbf{Preprocessing.}
Neural activity was standardized neuron-wise using statistics derived from the corresponding training data. Movie frames were centrally cropped from $36\times64$ to $36\times36$ pixels and resized to $192\times192$ (Fig.~\ref{fig:ca2real-overview}c). Exact normalization, synthetic-response filtering, and preprocessing procedures are provided in the Supplementary Methods.

\subsection*{Ca2REAL architecture and training}

Ca2REAL converts calcium activity into two neural-derived controls that condition a pretrained Wan2.1-VACE video generator \citep{wan2025,jiang2025} (Fig.~\ref{fig:ca2real-overview}d). The layout pathway provides coarse spatiotemporal structure, whereas the semantic pathway provides stimulus-specific visual information. The pretrained generator supplies a natural-video prior for details that are not fully specified by the neural signal. Additional implementation and training details are provided in the Supplementary Methods.

\noindent\textbf{Multiscale temporal front end.} Both control pathways receive a four-band temporal representation of the calcium sequence before pathway-specific encoding. We use a Haar-anchored adaptive lifting transform as the temporal front end to decompose calcium signals across multiple temporal scales, facilitating the extraction of features with distinct temporal dynamics \citep{sweldens1998,jia2022wavelets}.

\noindent\textbf{Layout control.}
The layout pathway predicts a low-resolution visual scaffold rather than a complete image, providing a coarse spatial constraint while leaving fine-scale detail to the video prior. A temporal convolutional encoder feeds a progressive multiresolution spatial decoder, with the temporal bands modulating each synthesis stage through residual FiLM-style affine conditioning. The output is a 60-frame, $64\times64$ grayscale movie that is upsampled to the VACE input resolution (Fig.~\ref{fig:ca2real-overview}d). This control constrains coarse geometry, luminance organization, object position, and temporal evolution while leaving high-frequency detail to the video prior. The layout controller was pretrained against low-resolution target movies.

\noindent\textbf{Semantic control.}
The semantic pathway uses an independently parameterized temporal encoder followed by a temporal Transformer \citep{vaswani2017} to represent stimulus-specific visual content. It predicts one global and eight temporally ordered 512-dimensional representations defined by a frozen CLIP ViT-B/32 image encoder \citep{radford2021}. The global token represents clip-level content, and the ordered tokens retain coarse changes in visual content across the clip. The semantic controller was trained against global and frame-level targets.

A learned token adapter projects the nine neural semantic representations into the 4,096-dimensional Wan text-conditioning space. These neural-derived tokens replace padded positions in the fixed UMT5 conditioning sequence, so stimulus-specific semantic information is provided by the calcium activity rather than by sample-specific language. In parallel, the predicted layout movie is supplied through the numerical VACE control pathway (Fig.~\ref{fig:ca2real-overview}d).

\noindent\textbf{Video-diffusion conditioning and joint training.}
 Wan2.1-VACE integrates the two neural controls and supplies the pretrained video prior. The Wan VAE, UMT5 encoder, original Wan transformer, original VACE parameters, and the lifting transforms and temporal encoders in both neural controllers were frozen during joint adaptation. Rank-16 LoRA modules \citep{hu2022} were trained in selected VACE control-block projections together with the semantic adapter and the task-specific layout and semantic decoder heads.

Joint optimization combined the VACE flow-matching objective \citep{lipman2023} with direct supervision of the two neural controls and a matched-versus-shuffled neural-dependence loss. For a clean video latent $\mathbf{y}$, Gaussian noise $\boldsymbol{\epsilon}$, and $u\sim U(0,1)$, we used
\begin{align}
\sigma
&=
\frac{3u}{1+2u},
&
\mathbf{y}_{\sigma}
&=
(1-\sigma)\mathbf{y}
+
\sigma\boldsymbol{\epsilon},
\\
\mathbf{v}^{*}
&=
\boldsymbol{\epsilon}-\mathbf{y},
&
\mathcal{L}_{\mathrm{VACE}}
&=
\left\|
\mathbf{v}_{\theta}
-
\mathbf{v}^{*}
\right\|_{2}^{2}.
\end{align}

The layout and semantic controls retained lightweight supervised anchors during joint adaptation:
\begin{align}
\mathcal{L}_{\mathrm{layout}}
&=
\mathcal{L}_{1}
+
0.1(1-\rho),
\\
\mathcal{L}_{\mathrm{semantic}}
&=
1-\overline{\cos}
+
0.25\,\mathcal{L}_{\Delta}.
\end{align}
Here, $\rho$ is the Pearson correlation between the predicted and target layout movies after flattening, $\overline{\cos}$ is the mean cosine similarity between matched predicted and target semantic tokens, and $\mathcal{L}_{\Delta}$ is the mean-squared error between successive frame-level semantic differences.

The complete joint objective was
\begin{align}
\mathcal{L}_{\mathrm{joint}}
={}&
\mathcal{L}_{\mathrm{VACE}}
+
0.10\,\mathcal{L}_{\mathrm{layout}}
\nonumber\\
&+
0.05\,\mathcal{L}_{\mathrm{semantic}}
+
0.50\,\mathcal{L}_{\mathrm{pair}}.
\end{align}

To encourage the generator to use the neural condition rather than rely only on the pretrained video prior, we included a matched-versus-shuffled dependence loss. Using the same target latent, diffusion state, and noise realization, we define
\begin{equation}
\Delta_{\mathrm{pair}}
=
\mathcal{L}_{\mathrm{shuffled}}
-
\mathcal{L}_{\mathrm{matched}},
\end{equation}
and
\begin{equation}
\mathcal{L}_{\mathrm{pair}}
=
\max\left(0,\,0.1-\Delta_{\mathrm{pair}}\right).
\label{eq:pair-dependence}
\end{equation}
This encourages the matched neural condition to explain the target better than a mismatched condition by a margin of 0.1.

\subsection*{Baselines and evaluation}

\noindent\textbf{Baseline decoders.}
Ca2REAL was compared with Ridge regression \citep{hoerl1970}, SID \citep{zhang2020sid}, and WISA \citep{peng2025wisa} using identical population-specific training and test memberships. Ridge regularization was selected by five-fold cross-validation restricted to the training set and the model was subsequently refitted on all training samples. SID and WISA used the maximum training horizons of their original configurations, and their final-epoch checkpoints were used for evaluation. Full baseline architectures and optimization settings are provided in the Supplementary Methods.

\noindent\textbf{Inference and photometric calibration.}
All Ca2REAL test clips were reconstructed with a fixed video-diffusion sampling protocol. To remove method-specific global luminance and contrast offsets without using test information, a separate affine photometric transformation was fitted for each decoder using the same fixed set of 128 training clips and was frozen before test evaluation. Main-text quantitative comparisons in Figs.~\ref{fig:temporary-evaluation-sketch} and \ref{fig:temporary-digital-twin-sketch} use these calibrated outputs; the corresponding uncalibrated reconstruction metrics are reported in Supplementary Tables 1 and 2. Qualitative examples use the same train-derived calibration.

\noindent\textbf{Evaluation metrics.} We evaluated complementary aspects of reconstruction rather than treating a single score as reconstruction accuracy (Table~\ref{tab:evaluation-metrics}). MSE, PSNR, SSIM, and MS-SSIM quantify pixel or local structural agreement; LPIPS, DISTS, and DreamSim quantify perceptual similarity \citep{wang2004ssim,wang2003msssim,zhang2018lpips,ding2022dists,fu2023dreamsim}; and DINOv2 and SigLIP2 compare higher-level visual representations \citep{oquab2024dinov2,tschannen2025siglip2}. Exact implementations, feature extractors, and frame-sampling procedures are provided in the Supplementary Methods.

\begin{table}[!t]
\centering
\caption{\textbf{Reconstruction metrics.}
Arrows indicate whether higher ($\uparrow$) or lower ($\downarrow$) values indicate better reconstruction.}
\label{tab:evaluation-metrics}
\scriptsize
\setlength{\tabcolsep}{2.5pt}
\renewcommand{\arraystretch}{1.02}

\begin{tabular}{@{}p{0.15\textwidth} p{0.34\textwidth} p{0.45\textwidth}@{}}
\toprule
\textbf{Metric} & \textbf{Formula} & \textbf{Meaning and notation} \\
\midrule

MSE $\downarrow$
&
$\displaystyle
\frac{1}{P}
\left\|
\mathbf{R}-\mathbf{T}
\right\|_2^2
$
&
Mean pixel-wise error between reconstruction $\mathbf R$ and target $\mathbf T$; $P$ is the total number of evaluated pixels. \\

PSNR $\uparrow$
&
$\displaystyle
10\log_{10}
\left(
\frac{L^2}{\mathrm{MSE}}
\right)
$
&
Logarithmic pixel fidelity; $L$ is the maximum pixel value. \\

SSIM $\uparrow$
&
\resizebox{0.98\linewidth}{!}{$\displaystyle
\frac{
(2\mu_R\mu_T+C_1)(2\sigma_{RT}+C_2)
}{
(\mu_R^2+\mu_T^2+C_1)
(\sigma_R^2+\sigma_T^2+C_2)
}
$}
&
Local structural similarity; $\mu_R,\mu_T$ are local means, $\sigma_R^2,\sigma_T^2$ are local variances, $\sigma_{RT}$ is local covariance, and $C_1,C_2$ are stabilization constants. \\

MS-SSIM $\uparrow$
&
$\displaystyle
l_M^{\alpha_M}
\prod_{m=1}^{M}
c_m^{\beta_m}s_m^{\gamma_m}
$
&
Multiscale structural similarity; $m$ indexes scale, $M$ is the coarsest scale, and $l_M$, $c_m$, and $s_m$ denote luminance, contrast, and structure terms weighted by $\alpha_M$, $\beta_m$, and $\gamma_m$. \\

LPIPS $\downarrow$
&
\resizebox{0.98\linewidth}{!}{$\displaystyle
\sum_l
\frac{1}{H_lW_l}
\sum_{h,w}
\left\|
\mathbf w_l\odot
\bigl(
\hat{\phi}_l(\mathbf R)_{hw}
-
\hat{\phi}_l(\mathbf T)_{hw}
\bigr)
\right\|_2^2
$}
&
Deep-feature perceptual distance; $l$ indexes feature layers, $\hat{\phi}_l$ denotes channel-normalized features, $H_l,W_l$ are their spatial dimensions, $(h,w)$ is spatial position, $\mathbf w_l$ contains learned channel weights, and $\odot$ denotes element-wise multiplication. \\

DISTS $\downarrow$
&
$\displaystyle
1-
\sum_{l,c}
\left(
\alpha_{lc}S_{lc}
+
\beta_{lc}T_{lc}
\right)
$
&
Deep-feature structural-textural distance; $l$ and $c$ index feature layer and channel, $S_{lc}$ and $T_{lc}$ are structural and texture similarity terms, and $\alpha_{lc},\beta_{lc}$ are learned weights. \\

DreamSim $\downarrow$
&
\resizebox{0.98\linewidth}{!}{$\displaystyle
\frac{1}{K}
\sum_{k=1}^{K}
\left[
1-
\cos
\bigl(
\phi(\mathbf R_k),
\phi(\mathbf T_k)
\bigr)
\right]
$}
&
Mean frame-wise perceptual distance; $K$ is the number of frames, $k$ indexes frames, $\mathbf R_k,\mathbf T_k$ are reconstructed and target frames, $\phi$ is the DreamSim embedding. \\

DINOv2 $\uparrow$
&
\resizebox{0.98\linewidth}{!}{$\displaystyle
\cos
\left(
\overline{\phi}_{\mathrm{DINO}}(\mathbf R),
\overline{\phi}_{\mathrm{DINO}}(\mathbf T)
\right)
$}
&
Similarity of high-level visual representations; $\phi_{\mathrm{DINO}}$ is the DINOv2 frame embedding, $\overline{\phi}_{\mathrm{DINO}}$ denotes temporal averaging over frames. \\

SigLIP2 $\uparrow$
&
\resizebox{0.98\linewidth}{!}{$\displaystyle
\cos
\left(
\overline{\phi}_{\mathrm{SigLIP2}}(\mathbf R),
\overline{\phi}_{\mathrm{SigLIP2}}(\mathbf T)
\right)
$}
&
Similarity of visual-semantic representations; $\phi_{\mathrm{SigLIP2}}$ is the SigLIP2 frame embedding, $\overline{\phi}_{\mathrm{SigLIP2}}$ denotes temporal averaging over frames. \\

\bottomrule
\end{tabular}
\end{table}

\noindent\textbf{Stimulus-identification accuracy.}
To quantify stimulus-specific decoding, we used all-pairs stimulus-identification accuracy. Unlike a similarity score computed only for one matched reconstruction-target pair, this statistic asks whether the reconstructed clip is closer in SigLIP2 space to its own target than to each mismatched target. We extracted pooled, L2-normalized SigLIP2 representations for every reconstruction and target and computed their similarity by inner product. For reconstruction $i$, its similarity $S_{ii}$ to the matched target was compared with its similarity $S_{ij}$ to every mismatched target $j\neq i$. Pairwise identification credit was defined as
\begin{equation}
a_{ij}
=
\mathbb{I}(S_{ii}>S_{ij})
+
\frac{1}{2}\mathbb{I}(S_{ii}=S_{ij}),
\end{equation}
so a correct comparison receives 1, a tie 0.5, and an incorrect comparison 0. The reported all-pairs accuracy is the mean of $a_{ij}$ over all ordered matched-versus-mismatched comparisons. 

\noindent\textbf{Visual-content clusters.} To assess performance across heterogeneous natural scenes, we grouped the held-out targets into four visual-content clusters and evaluated performance within each group. Clustering used only frozen target-movie SigLIP2 representations and was independent of all model reconstructions and reconstruction scores. The four clusters were labeled post hoc according to their dominant visual content: \emph{Portrait}, containing primarily people, faces, and portrait- or indoor-dominated scenes; \emph{Landscape}, containing outdoor landscapes, roads, and travel scenes; \emph{Motion / Buildings}, containing prominent human motion, action, and building- or urban-dominated scenes; and \emph{Low visibility}, containing dark, low-contrast, texture-dominated, or visually ambiguous clips. 

\subsection*{Statistics and reproducibility}

Statistical analyses were performed on held-out test clips, with each clip treated as the unit of analysis. The measured V1 experiment contained 178 test clips from one mouse, and the virtual-population experiment contained 946 test clips. Confidence intervals were estimated by bootstrap resampling unless otherwise specified. When methods or perturbation conditions were evaluated on the same clips, paired comparisons were used and Holm correction was applied within each prespecified family of multiple comparisons. The donor-similarity analysis used Spearman correlation with a query-clustered permutation test, with confidence intervals obtained by bootstrap resampling at the query level. Exact procedures for individual analyses are described in the corresponding figure legends and Supplementary Methods.

\subsection*{Data Availability}

The Sensorium 2023 dataset is publicly available at \url{https://gin.g-node.org/pollytur/Sensorium2023Data} and \url{https://gin.g-node.org/pollytur/sensorium_2023_dataset}. MICrONS data used in this study are publicly available through the MICrONS Explorer at \url{https://www.microns-explorer.org/cortical-mm3}.

\subsection*{Code availability}

Code supporting the findings of this study is available from the corresponding author upon reasonable request.

\backmatter

\section*{Acknowledgements}

This work was supported by the Science and Technology Innovation 2030--Brain Science and Brain-Inspired Intelligence Project (2021ZD0200204), the National Natural Science Foundation of China (92570202), Shanghai Municipal Science and Technology (24JS2810400), and SIMIS (SIMISID-2025-NC). The computations in this research were performed using the CFFF platform of Fudan University.

\section*{Author contributions}

Y.Y. conceived the study. H.A.W., J.Z., and Y.Y. designed the study and developed the methodology. Y.H. implemented the simulation framework. Y.H., H.A.W., and J.Z. conducted the main experiments. J.M., J.D., X.C., Z.G., and Y.L. contributed to the experiments and data analysis. All authors contributed to figure preparation. Y.H., H.A.W., J.Z., and Y.Y. drafted and revised the manuscript. Y.Y. supervised the project. All authors reviewed and approved the manuscript.

\section*{Competing interests}

The authors declare no competing interests.

\end{document}